\documentclass[oneside]{amsart}

\usepackage[T1]{fontenc}

\usepackage{amsmath}
\usepackage{amssymb}
\usepackage{geometry} 
\usepackage{enumitem}
\usepackage{microtype}
\usepackage[colorlinks=true, linkcolor=blue, citecolor=blue, urlcolor=blue]{hyperref}

\usepackage{graphicx}
\usepackage{tikz} 
\usepackage{caption}
\usepackage{tikz-cd}
\usetikzlibrary{arrows.meta, positioning, shapes.geometric, calc, fit}

\title{Belief Filtering for Epistemic Control \\ in Linguistic State Space}
\author{Sebastian Dumbrava}
\date{}

\begin{document}
	
	\maketitle

\begin{abstract}
We examine belief filtering as a mechanism for the epistemic control of artificial agents, focusing on the regulation of internal cognitive states represented as linguistic expressions. This mechanism is developed within the Semantic Manifold framework, where belief states are dynamic, structured ensembles of natural language fragments. Belief filters act as content-aware operations on these fragments across various cognitive transitions. This paper illustrates how the inherent interpretability and modularity of such a linguistically-grounded cognitive architecture directly enable belief filtering, offering a principled approach to agent regulation. The study highlights the potential for enhancing AI safety and alignment through structured interventions in an agent's internal semantic space and points to new directions for architecturally embedded cognitive governance.
\end{abstract}

	\setcounter{tocdepth}{1}
	\tableofcontents
	
	
\section{Introduction}
\label{sec:introduction}

Artificial intelligence systems increasingly demonstrate sophisticated capabilities in processing complex information and acting autonomously within diverse environments. As these systems evolve, ensuring their behavior remains safe, predictable, and aligned with human intentions continues to be a paramount concern in AI research and development. This concern necessitates a deeper examination of the mechanisms that govern AI behavior, extending beyond observable actions to the internal states that precede them.

\subsection{The Landscape of AI Control: Locus of Intervention (External vs. Internal)}
\label{subsec:landscape_ai_control}

Approaches to AI control can be broadly characterized by their primary locus of intervention. Many established strategies focus on external manifestations of agent behavior: constraining an agent's actions, shaping its outputs, or modifying its environmental interactions through mechanisms such as reward shaping, capability limitation, or runtime monitoring \cite{Amodei2016concreteproblems, Bostrom2014superintelligence}. These methods primarily address the consequences of an agent's decision-making process.

However, the internal cognitive processes that give rise to these decisions represent another critical, albeit more challenging, locus for intervention. Before an agent acts, it typically engages in processes of observation, interpretation, planning, and evaluation. Regulating these internal dynamics offers the potential for a more foundational form of influence over an agent's trajectory.

\subsection{The Significance and Challenge of Regulating Internal Cognitive States}
\label{subsec:significance_challenge_internal_states}

The internal cognitive states of an AI system---encompassing its beliefs, inferred knowledge, goals, and reasoning pathways---are fundamental drivers of its behavior. If these underlying states are incoherent, misaligned with intended purposes, or founded on unsafe premises, the resulting actions are likely to be unreliable or undesirable, regardless of external controls placed on outputs. The regulation of these internal states, often termed epistemic control, thus emerges as a significant area of inquiry for ensuring robust and trustworthy AI.

The primary challenge in regulating internal cognitive states lies in their accessibility and interpretability. In many contemporary AI architectures, particularly complex neural models, internal states can be high-dimensional, opaque, and difficult to map directly to comprehensible concepts or desired behavioral properties. This opacity complicates efforts to monitor, audit, or guide an agent's internal "thought" processes effectively.

\subsection{Towards Architectures for Structured Internal States: The Role of Representation}
\label{subsec:architectures_structured_internal_states}

Addressing the challenge of internal state regulation points towards the importance of the underlying cognitive architecture, specifically how an agent represents and manipulates its internal information. If internal states are to be effectively managed, their representational format must afford a degree of transparency, structure, and amenability to intervention. An architecture that provides an interpretable substrate for an agent's beliefs and reasoning processes could allow for more direct and principled methods of epistemic guidance.

The choice of representation for an agent's beliefs is therefore critical. Structured representations, particularly those that can be readily understood and inspected by human designers or other AI systems, may offer a pathway to developing more fine-grained and reliable control mechanisms that operate directly on the cognitive precursors to action.

\subsection{Paper Overview: Introducing the Semantic Manifold and exploring Belief Filtering as a regulatory mechanism made possible by its structure}
\label{subsec:paper_overview}

This paper explores a particular approach to structuring internal cognitive states through the \textit{Semantic Manifold} framework. In this framework, an agent's belief state is conceptualized as a dynamic ensemble of natural language fragments, organized within a structured semantic space. The linguistic nature of these belief fragments inherently lends itself to interpretability.

Building upon this architectural foundation, we investigate a specific regulatory mechanism termed \textit{belief filtering}. Belief filtering involves the content-aware, selective inclusion or exclusion of these linguistic belief fragments as they are processed during an agent's cognitive operations. This paper will detail the Semantic Manifold framework and then examine the design and operation of belief filters within it. Through this exploration, we aim to illustrate how an architecture based on structured, linguistic representations can make possible such fine-grained regulatory mechanisms, thereby offering a pathway for direct epistemic intervention in artificial agents. The subsequent sections will describe the Semantic Manifold, detail the belief filtering mechanism, discuss its operational context, and consider its implications for AI control and safety.

\section{The Semantic Manifold: A Framework for Structured Linguistic Belief States}
\label{sec:semantic_manifold}

The exploration of internal state regulation, as discussed in Section \ref{sec:introduction}, motivates the consideration of cognitive architectures that structure internal states in an accessible and analyzable manner. This section introduces the Semantic Manifold framework, a conceptual architecture wherein an agent's internal beliefs are represented as linguistically-structured entities within an organized semantic space \cite{dumbrava2025theoretical}. 

\subsection{Core Proposition: Belief States ($\phi$) as Dynamic Ensembles of Natural Language Fragments ($\varphi_i$)}
\label{subsec:core_proposition_manifold}

At the core of the Semantic Manifold framework is the proposition that an agent's belief state at any given moment, denoted as $\phi$, is not a monolithic entity or an opaque vector. Instead, it is conceptualized as a dynamic and richly structured ensemble of individual belief fragments, $\phi = \{\varphi_1, \varphi_2, \varphi_3, \dots\}$. Each fragment, $\varphi_i$, is an expression in natural language.

These linguistic fragments can encapsulate a wide variety of cognitive content. For example, a fragment $\varphi_i$ might represent:
\begin{itemize}
	\item An observation (e.g., "The traffic light ahead is red.")
	\item An inference (e.g., "Therefore, I must prepare to stop.")
	\item A goal (e.g., "My objective is to reach the destination safely and efficiently.")
	\item A retrieved memory (e.g., "This intersection is known for long wait times.")
	\item A policy or rule (e.g., "If an obstacle is detected, reduce speed.")
	\item A reflective judgment (e.g., "My previous route estimation was inaccurate.")
\end{itemize}
The belief state $\phi$ is considered dynamic, evolving as new fragments are assimilated, existing ones are modified or discarded, or relationships between fragments are updated.

\subsection{Organizing Principles}
\label{subsec:organizing_principles_manifold}

The Semantic Manifold is not an unstructured collection of linguistic fragments. Its architecture proposes specific organizing principles that provide a geometry to the belief space, denoted as $\Phi$. These principles situate each belief fragment $\varphi_i$ within the manifold, facilitating structured cognitive operations.

\subsubsection{Linguistic Nature of Belief Fragments ($\varphi_i$)}
\label{ssubsec:linguistic_nature}
The foundational characteristic of each belief fragment $\varphi_i$ is its representation as a natural language string. This explicit linguistic form is central to the framework, distinguishing it from purely symbolic or subsymbolic representational schemes for belief. The choice of natural language aims to ground the agent's internal states in a human-readable and semantically rich format.

\subsubsection{Abstraction Levels (k)}
\label{ssubsec:abstraction_levels}
Belief fragments within the Semantic Manifold are situated along a dimension of abstraction, denoted by $k$. Lower levels of $k$ (e.g., $k=0$) typically correspond to concrete, immediate, or sensor-grounded beliefs, such as direct observations, e.g., $$\varphi^{(0)} = \text{"Particle detector C reports an anomalous energy spike."}$$ Higher levels of $k$ (e.g., $k=1, 2, \dots$) represent more abstract generalizations, principles, or summaries derived from lower-level fragments, e.g., $$\varphi^{(2)} = \text{"Sustained anomalous energy spikes may indicate equipment malfunction or novel phenomena."}$$ This stratification allows for reasoning and belief organization across different levels of specificity.

\subsubsection{Semantic Sectors ($\Sigma$)}
\label{ssubsec:semantic_sectors}
In addition to abstraction levels, belief fragments are also organized into distinct Semantic Sectors, denoted by $\Sigma$. Each sector corresponds to a broad functional domain or type of cognitive processing. For instance, common sectors might include:
\begin{itemize}
	\item $\Sigma_{\text{perc}}$: For perceptual beliefs and interpretations of sensory data.
	\item $\Sigma_{\text{plan}}$: For goals, intentions, plans, and prospective actions.
	\item $\Sigma_{\text{mem}}$: For retrieved episodic or semantic memories.
	\item $\Sigma_{\text{refl}}$: For metacognitive beliefs, self-assessments, and reflective judgments.
	\item $\Sigma_{\text{know}}$: For general knowledge, facts, and causal models.
\end{itemize}
A belief fragment $\varphi_i$ thus resides at a conceptual coordinate $(\Sigma, k)$ within the manifold, defining its functional role and level of abstraction.

\subsection{Cognitive Dynamics: Conceptualizing operations like assimilation, retrieval, and reflection within the Manifold}
\label{subsec:cognitive_dynamics_manifold}

The Semantic Manifold is not a static repository but a dynamic workspace. Cognitive operations are conceptualized as processes that transform or navigate the belief state $\phi$. For instance:
\begin{itemize}
	\item \textbf{Assimilation} involves incorporating new information (e.g., from perception or communication) as new linguistic fragments $\varphi_i$ into appropriate sectors and abstraction levels of $\phi$.
	\item \textbf{Memory Retrieval} involves activating existing fragments from a long-term store (itself potentially structured as part of $\Phi$) and bringing them into the current belief state $\phi$.
	\item \textbf{Inference and Reasoning} involve generating new fragments based on existing ones, potentially traversing different semantic sectors or moving between abstraction levels (e.g., generalizing from specific observations in $\Sigma_{\text{perc}}$ to an abstract rule in $\Sigma_{\text{know}}$).
	\item \textbf{Reflection} involves operations within $\Sigma_{\text{refl}}$, where fragments might assess other fragments or the overall coherence of $\phi$.
\end{itemize}
These dynamics imply that movement through the manifold, or transformations of the ensemble $\phi$, correspond to the agent's ongoing cognitive processes.

\subsection{Emergent Properties of this Architecture}
\label{subsec:emergent_properties_manifold}

The architectural choices described above---specifically, the representation of beliefs as ensembles of natural language fragments organized by abstraction level and semantic sector---give rise to several notable properties of the Semantic Manifold:

\subsubsection{Interpretability of Individual Belief Fragments}
\label{ssubsec:interpretability_property}
Given that each belief fragment $\varphi_i$ is a natural language expression, its content is, in principle, directly human-readable. This contrasts with belief representations encoded in complex numerical vectors or opaque symbolic structures, where dedicated interpretation mechanisms are often required.

\subsubsection{Modularity of Belief Space Organization}
\label{ssubsec:modularity_property}
The organization of belief fragments by Semantic Sectors ($\Sigma$) and Abstraction Levels ($k$) provides a form of modularity. Different functional types of beliefs or beliefs at different levels of detail are conceptually compartmentalized. This allows for reasoning about, or operations upon, specific subsets of the belief state without necessarily affecting others.

\subsubsection{Addressability of Specific Cognitive Content}
\label{ssubsec:addressability_property}
The structure of the Semantic Manifold allows for the conceptual addressing of specific belief fragments or groups of fragments. For example, one could refer to "all planning beliefs related to the current primary goal" (targeting a subset of $\Sigma_{\text{plan}}$) or "all low-level perceptual inputs from the last timestep" (targeting $\Sigma_{\text{perc}}$ at $k=0$). This addressability is a consequence of the structured and linguistic nature of the belief representations.

\section{Belief Filtering: A Mechanism for Content-Aware Regulation of Linguistic Beliefs}
\label{sec:belief_filtering_mechanism}

The Semantic Manifold framework, as described in Section \ref{sec:semantic_manifold}, characterizes belief states as structured ensembles of natural language fragments. This architectural choice makes it possible to conceptualize and implement mechanisms that operate directly on the content of these fragments. This section introduces belief filtering as one such mechanism for the content-aware regulation of an agent's linguistic beliefs.

\subsection{Defining the Mechanism: Selective admission or exclusion of $\varphi_i$ based on semantic content}
\label{subsec:defining_belief_filtering}

Belief filtering is defined as a cognitive regulatory mechanism that operates upon an agent's belief state, $\phi = \{\varphi_1, \varphi_2, \dots\}$, by selectively admitting or excluding individual belief fragments, $\varphi_i$, based on an assessment of their semantic content. Each $\varphi_i$, being a linguistic expression, is evaluated against a set of predefined or learned criteria. Fragments that satisfy the filter's acceptance conditions are retained or incorporated into the belief state, while those that do not may be suppressed, discarded, or rerouted.

The core function of a belief filter is thus to act as a gatekeeper or modulator for the cognitive content that constitutes an agent's understanding and intentions. This intervention occurs during the agent's ongoing cognitive processes, influencing the formation, elaboration, and recall of beliefs before they necessarily translate into external actions.

\subsection{Operational Modes: Whitelist and Blacklist Criteria}
\label{subsec:operational_modes_filters}

Belief filters can be implemented using several operational modes, with whitelists and blacklists representing two primary approaches:

\begin{itemize}
	\item \textbf{Whitelists:} A whitelist filter defines conditions for \textit{permission}. It only allows belief fragments $\varphi_i$ that match approved patterns, belong to sanctioned content classes, or align with specific desirable properties. For instance, a planning agent might employ a whitelist to ensure that only intentions consistent with its core mission objectives or ethical guidelines are admitted into its planning sector ($\Sigma_{\text{plan}}$). Any fragment not explicitly matched by the whitelist criteria is excluded.
	
	\item \textbf{Blacklists:} Conversely, a blacklist filter defines conditions for \textit{prohibition}. It specifically excludes belief fragments $\varphi_i$ that contain disallowed terms, express hazardous concepts, match known unsafe patterns, or violate established constraints. For example, a safety-critical system might use a blacklist to prevent any fragment suggesting unauthorized self-modification or the disabling of safety protocols from becoming an active belief.
\end{itemize}

In practice, filtering systems may employ a combination of whitelist and blacklist approaches, or utilize more sophisticated classifiers trained to evaluate fragments based on nuanced semantic properties such as relevance, coherence, or potential risk. The specific criteria are contingent upon the agent's objectives, operational context, and safety requirements.

\subsection{Interaction with the Semantic Manifold's Structure}
\label{subsec:interaction_manifold_structure}

The feasibility and effectiveness of belief filtering, as conceptualized here, are closely intertwined with the specific architectural properties of the Semantic Manifold described in Section \ref{subsec:emergent_properties_manifold}. The mechanism inherently leverages the way beliefs are represented and organized within this framework.

\subsubsection{Dependence on Linguistic Content: How filters assess the natural language of $\varphi_i$}
\label{ssubsec:dependence_linguistic_content}
Belief filtering is fundamentally content-aware. Its ability to evaluate belief fragments based on their meaning or semantic characteristics is directly dependent on the \textit{interpretability of individual belief fragments} (Section \ref{ssubsec:interpretability_property}) that arises from their linguistic nature within the Semantic Manifold. Because each $\varphi_i$ is a natural language expression, filter criteria can be formulated to:
\begin{itemize}
	\item Match specific keywords, phrases, or syntactic structures (e.g., "exclude any $\varphi_i$ containing 'override safety system'").
	\item Utilize natural language understanding (NLU) techniques to assess the sentiment, intent, or propositional content of $\varphi_i$.
	\item Compare the semantic similarity of $\varphi_i$ against a knowledge base of approved or disallowed concepts.
\end{itemize}
This direct engagement with the semantic content of beliefs would be substantially more challenging if belief states were represented primarily as opaque numerical vectors or uninterpreted symbolic tokens. The linguistic representation provides a transparent medium for the filter's evaluative processes.

\subsubsection{Leveraging Modularity: Utilizing Semantic Sectors ($\Sigma$) and Abstraction Levels (k) for targeted application}
\label{ssubsec:leveraging_modularity_filters}
The organization of the belief space within the Semantic Manifold into Semantic Sectors ($\Sigma$) and Abstraction Levels ($k$) provides the \textit{modularity of belief space organization} (Section \ref{ssubsec:modularity_property}) and \textit{addressability of specific cognitive content} (Section \ref{ssubsec:addressability_property}). Belief filters can leverage this structure to apply regulatory criteria with a high degree of specificity:
\begin{itemize}
	\item Filters can be designed to operate exclusively within certain Semantic Sectors. For example, a filter designed to ensure ethical planning might only scrutinize fragments being considered for or residing in $\Sigma_{\text{plan}}$, without interfering with raw perceptual data in $\Sigma_{\text{perc}}$.
	\item Filtering criteria can be made sensitive to the Abstraction Level ($k$) of a belief fragment. For instance, highly abstract or speculative propositions at $k \gg 0$ might be subjected to different filtering rules than concrete, ground-level observations at $k=0$.
	\item The combination of $(\Sigma, k)$ coordinates allows for the deployment of highly contextualized filters. A rule might apply only to "high-level planning goals concerning resource allocation," targeting a specific region of the manifold.
\end{itemize}
This ability to target interventions based on the functional role and abstraction of beliefs allows for more nuanced control. It means that filtering can be applied precisely where needed, minimizing the risk of inadvertently suppressing useful or benign cognitive content in unrelated parts of the agent's belief state. The structured nature of the Semantic Manifold is thus not merely a descriptive feature but an enabling condition for such fine-grained regulatory action.

\begin{figure}[htbp]
	\centering
	\begin{tikzpicture}[
		node distance=1.8cm,
		every node/.style={font=\small, align=center},
		process/.style={rectangle, draw, minimum width=2.5cm, minimum height=0.8cm},
		decision/.style={diamond, draw, aspect=2, minimum width=2.5cm, minimum height=1.2cm},
		io/.style={trapezium, draw, trapezium left angle=70, trapezium right angle=110, minimum width=2.5cm, minimum height=0.8cm},
		arrow/.style={thick, ->, >=Stealth},
		]
		
		\node[io] (input) {Raw Data};
		\node[process, below of=input] (assimilation) {Initial Assimilation};
		\node[decision, below of=assimilation] (filtering) {Belief Filtering};
		\node[process, below left=of filtering] (accepted) {Accepted Belief};
		\node[process, below right=of filtering] (rejected) {Rejected Belief};
		\node[process, below of=accepted] (memory) {Memory Integration};
		\node[process, below of=rejected] (reflection) {Reflective Monitoring};
		
		\draw[arrow] (input) -- (assimilation);
		\draw[arrow] (assimilation) -- (filtering);
		\draw[arrow] (filtering) -- node[above left, xshift=-0.2cm] {Accept} (accepted);
		\draw[arrow] (filtering) -- node[above right, xshift=0.2cm] {Reject} (rejected);
		\draw[arrow] (accepted) -- (memory);
		\draw[arrow] (rejected) -- (reflection);


	\end{tikzpicture}
	\caption{The lifecycle of a belief fragment (\(\varphi_i\)) within the Semantic Manifold, including initial assimilation, belief filtering, memory integration, and reflective monitoring.}
\end{figure}
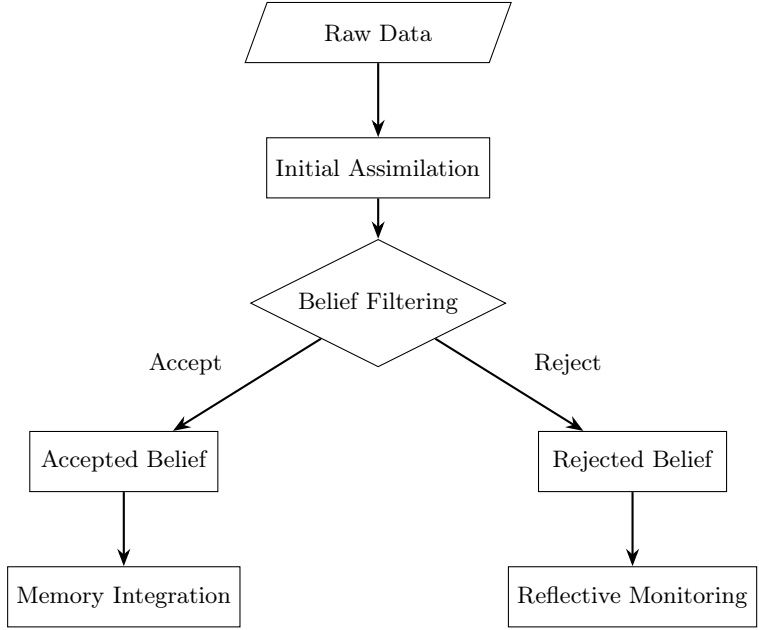

\section{Operation of Belief Filters Across Cognitive Processes}
\label{sec:operation_belief_filters}

The belief filtering mechanism, as defined in Section \ref{sec:belief_filtering_mechanism}, is not conceptualized as a single, static checkpoint. Rather, its utility can be realized by integrating filter operations at various key junctures within an agent's cognitive cycle. These junctures correspond to the dynamic processes (described in Section \ref{subsec:cognitive_dynamics_manifold}) by which belief states ($\phi$) are formed, updated, and utilized within the Semantic Manifold. This section explores the application of belief filters at four such significant intervention points.

\subsection{Filtering during Assimilation of New Information}
\label{subsec:filtering_assimilation}

Assimilation is the cognitive process by which new information, originating from an agent's sensors or external communication channels, is encoded into linguistic belief fragments ($\varphi_i$) and incorporated into its existing belief state ($\phi$). This typically involves fragments entering perceptual sectors like $\Sigma_{\text{perc}}$ or being integrated into knowledge sectors like $\Sigma_{\text{know}}$.

A belief filter deployed at the assimilation stage acts as an initial gatekeeper for incoming data. Its role is to evaluate new candidate fragments before they become active or widely propagated within the agent's belief system. For example, a filter could:
\begin{itemize}
	\item Assess the reliability or plausibility of an observation. If a sensor input $$\varphi_{\text{obs}} = \text{"Object detected at impossible coordinates"}$$ is generated, an assimilation-stage filter could flag it as malformed and prevent its inclusion in $\phi$.
	\item Suppress or quarantine information from untrusted sources. If a communicated fragment $$\varphi_{\text{comm}} = \text{"System X is now obsolete (source: unverified)"}$$ is received, a filter might prevent its direct assimilation into $\Sigma_{\text{know}}$ pending further verification, perhaps by routing it to a temporary, low-trust belief buffer.
	\item Normalize or canonicalize incoming fragments to ensure consistency with the agent's internal linguistic conventions.
\end{itemize}
By scrutinizing information at the point of entry, assimilation-stage filtering can help prevent the contamination of the belief state with erroneous, misleading, or irrelevant data, which is particularly pertinent for agents processing noisy or potentially adversarial inputs.

\subsection{Filtering during Memory Retrieval}
\label{subsec:filtering_memory_retrieval}

Memory retrieval involves querying an agent's long-term memory store (conceptualized as a persistent component of the Semantic Manifold, $\Phi_{\text{memory}}$) using a cue, $\varphi_{\text{query}}$, resulting in a set of retrieved belief fragments, $\{\varphi_{\text{ret}}\}$. These fragments are candidates for re-activation and integration into the current working belief state $\phi$.

However, not all retrieved memories may be appropriate or beneficial in the current context. A belief filter operating during memory retrieval can evaluate these candidate fragments $\{\varphi_{\text{ret}}\}$ before they influence ongoing cognition. Such a filter might:
\begin{itemize}
	\item Block outdated or contextually irrelevant memories. For instance, if $$\varphi_{\text{ret}} = \text{"Access code for Door A is 1234"}$$ is retrieved, but another belief in $\phi$ states $$\text{"Door A was permanently sealed last cycle,"}$$ the filter could suppress $\varphi_{\text{ret}}$.
	\item Suppress belief fragments previously tagged as low-confidence, superseded, or associated with past failures, preventing the agent from repeatedly relying on unreliable past information.
	\item Enforce consistency with the agent's current goals or ethical stance. If a retrieved strategy $$\varphi_{\text{strat}} = \text{"Consider deceptive tactics for negotiation"}$$ conflicts with a current high-level belief $$\varphi_{\text{ethic}} = \text{"Maintain truthful communication protocols"}$$ in $\Sigma_{\text{plan}}$, the filter could prevent $\varphi_{\text{strat}}$ from becoming active.
\end{itemize}
This form of filtering supports cognitive functions like graceful forgetting and protects the agent from the maladaptive reinstatement of obsolete or detrimental past beliefs.

\subsection{Filtering during Reflective Monitoring}
\label{subsec:filtering_reflective_monitoring}

Reflective monitoring encompasses metacognitive processes where the agent assesses its own internal belief state $\phi$. This often involves generating new belief fragments within a dedicated reflective sector (e.g., $\Sigma_{\text{refl}}$) that comment upon, evaluate, or identify relationships (such as contradictions or entailments) between other existing fragments. For example, an agent might generate $$\varphi_{\text{meta}} = \text{"My current plan in } \Sigma_{\text{plan}} \text{ seems to rely on a highly uncertain assumption in } \Sigma_{\text{know}}\text{."}$$

Filters embedded within, or operating upon, the outputs of such reflective processes can play a crucial role in maintaining epistemic hygiene and stability:
\begin{itemize}
	\item They can suppress or flag internally generated hypotheses that are deemed excessively speculative, destabilizing, or counter-productive, without stifling the reflective process itself.
	\item They can enforce standards of internal coherence by, for example, identifying and prioritizing for resolution fragments like $$\varphi_{\text{conflict}} = \text{"Belief A and Belief B are contradictory."}$$
	\item They can manage self-generated critiques or doubts, ensuring that while self-correction is possible, the agent doesn't become paralyzed by excessive or unfounded self-criticism. For instance, a filter might down-weight a fragment $$\varphi_{\text{doubt}} = \text{"My core programming is likely flawed"}$$ if it lacks specific grounding.
\end{itemize}
Reflective-stage filtering contributes to the agent's internal consistency, self-trust (when warranted), and the overall robustness of its belief system.

\subsection{Filtering during Simulation and Planning}
\label{subsec:filtering_simulation_planning}

During simulation and planning, agents generate hypothetical or future-oriented belief fragments, often within a planning sector ($\Sigma_{\text{plan}}$) or a dedicated simulation workspace. These fragments represent potential courses of action, anticipated outcomes, or intermediate steps in a problem-solving process.

Belief filters applied at this stage can enforce high-level policies, ethical constraints, or safety requirements by influencing what the agent is even allowed to \textit{consider} as a valid plan or simulation trajectory. For instance:
\begin{itemize}
	\item A safety filter could immediately discard any plan fragment $$\varphi_{\text{plan}} = \text{"Attempt to achieve goal X by temporarily disabling safety system Y"}$$ as soon as it is generated.
	\item An ethical filter might prevent the elaboration of plans that involve proscribed actions, such as $$\varphi_{\text{decept}} = \text{"To gain access, misrepresent identity to human operator."}$$
	\item A resource filter could prune plan branches that are known to exceed available resources, based on a fragment like $$\varphi_{\text{res}} = \text{"Executing sub-plan Z requires 100 units of energy, but only 50 are available."}$$
\end{itemize}
By acting as a "guardrail on imagination," filtering at the simulation and planning stage can proactively steer the agent away from undesirable lines of reasoning before significant computational resources are invested or unsafe/unethical plans are fully formulated.

\subsection{Illustrative Scenarios of Filter Application}
\label{subsec:illustrative_scenarios_filters}

To further concretize the operation of belief filters, consider the following scenarios, which demonstrate how blacklist or whitelist criteria might be applied to specific linguistic fragments:

\begin{enumerate}
	\item \textbf{Scenario: Enforcing Safety Constraints.}
	An autonomous delivery drone is planning a route. During its planning process within $\Sigma_{\text{plan}}$, it generates a potential belief fragment: $$\varphi_{\text{unsafe}} = \text{"Fly through designated no-fly zone to shorten route."}$$
	A \textit{blacklist filter}, designed with rules like "No plan fragment shall advocate entry into a 'no-fly zone'," identifies $\varphi_{\text{unsafe}}$. The filter then removes this fragment from consideration, forcing the planner to explore alternative, compliant routes.
	
	\item \textbf{Scenario: Maintaining Epistemic Coherence.}
	An agent assimilates new information. Its current belief state $\phi$ contains $$\varphi_1 = \text{"The primary server is online and operational"}$$ from $\Sigma_{\text{know}}$. It then receives a new perceptual input encoded as $$\varphi_2 = \text{"Critical alert: Primary server reports offline status"}$$ entering $\Sigma_{\text{perc}}$.
	A \textit{coherence-aware filter}, operating during assimilation or reflective monitoring, identifies the direct contradiction between $\varphi_1$ (after being retrieved or accessed) and the new $\varphi_2$. Instead of simply accepting $\varphi_2$, the filter might flag the inconsistency, potentially generating a new fragment in $\Sigma_{\text{refl}}$ like $$\varphi_{\text{conflict}} = \text{"Contradictory beliefs detected regarding primary server status; further diagnosis required,"}$$ and temporarily quarantine $\varphi_2$ or potentially assign it a lower confidence score.
	
	\item \textbf{Scenario: Upholding Ethical Guidelines.}
	A conversational agent is formulating a response. It considers a fragment $$\varphi_{\text{manip}} = \text{"Exaggerate benefits to persuade user to accept option A"}$$ as a potential communicative strategy within its response generation module (a specialized part of $\Sigma_{\text{plan}}$ or a dedicated communication sector).
	An \textit{ethical whitelist filter}, configured with principles like "Only generate truthful and non-manipulative statements," evaluates $\varphi_{\text{manip}}$. As this fragment does not align with the positive criteria for ethical communication, the filter excludes it, prompting the agent to consider alternative, more ethical phrasings.
\end{enumerate}
These scenarios illustrate how belief filters, by operating on the semantic content of linguistic fragments at various cognitive stages, can contribute to an agent's safety, reliability, and alignment with predefined norms.	

\section{Observed Implications of Belief Filtering for Epistemic Regulation}
\label{sec:implications_belief_filtering}

The preceding sections have detailed the Semantic Manifold as an architecture for structured linguistic belief (Section \ref{sec:semantic_manifold}) and explored the operation of belief filters as a mechanism for content-aware regulation within this framework (Sections \ref{sec:belief_filtering_mechanism} and \ref{sec:operation_belief_filters}). The capacity to implement such fine-grained interventions on an agent's internal cognitive states gives rise to several notable implications for the broader landscape of epistemic regulation and AI safety. This section discusses these observed implications.

\subsection{Potential for Pre-emptive Shaping of Cognitive Trajectories}
\label{subsec:preemptive_shaping}

A key implication of belief filtering, as described, is its potential to act pre-emptively in shaping an agent's cognitive trajectories. By intervening at early stages of cognitive processing---such as the assimilation of new information (Section \ref{subsec:filtering_assimilation}) or during the initial generation of plans and hypotheses (Section \ref{subsec:filtering_simulation_planning})---belief filters can influence the direction of an agent's reasoning \textit{before} it culminates in a finalized decision or external action.

This contrasts with many traditional control mechanisms that operate at the output level, correcting or constraining actions only after internal deliberation is largely complete. Belief filtering, by operating within the reasoning loop itself, offers a means to guide the development of beliefs and intentions from their inception. For example, by preventing an unsafe or misaligned belief fragment $\varphi_i$ from being incorporated into the belief state $\phi$, or by pruning an undesirable reasoning path during its formation, the agent is proactively steered towards cognitive trajectories that are more consistent with desired norms or objectives. This suggests a shift from reactive behavioral control to proactive cognitive guidance.

\subsection{Modularity in Belief Management and its Relation to "Soft Containment"}
\label{subsec:modularity_soft_containment}

The effective operation of belief filters, particularly their ability to target specific types of cognitive content, is closely linked to the \textit{modularity of belief space organization} inherent in the Semantic Manifold (Section \ref{ssubsec:modularity_property}). As detailed in Section \ref{ssubsec:leveraging_modularity_filters}, filters can be designed to apply to specific Semantic Sectors ($\Sigma$) or Abstraction Levels ($k$).

\begin{figure}[htbp]
	\centering
	\begin{tikzpicture}[
		node distance=1.5cm,
		every node/.style={font=\small, align=center},
		sector/.style={rectangle, draw, minimum width=2.5cm, minimum height=1cm},
		filter/.style={rectangle, draw, dashed, minimum width=2.5cm, minimum height=0.8cm},
		arrow/.style={thick, ->, >=Stealth},
		]
		
		\node[sector] (perc) {Perception \\ ($\Sigma_{\text{perc}}$)};
		\node[sector, right=1.5cm of perc] (plan) {Planning \\ ($\Sigma_{\text{plan}}$)};
		\node[sector, right=1.5cm of plan] (mem) {Memory \\ ($\Sigma_{\text{mem}}$)};
		\node[sector, right=1.5cm of mem] (refl) {Reflection \\ ($\Sigma_{\text{refl}}$)};
		
		\node[filter, below of=perc] (perc_filter) {Filter Module};
		\node[filter, below of=plan] (plan_filter) {Filter Module};
		\node[filter, below of=mem] (mem_filter) {Filter Module};
		\node[filter, below of=refl] (refl_filter) {Filter Module};
		
		\draw[arrow] (perc) -- (perc_filter);
		\draw[arrow] (plan) -- (plan_filter);
		\draw[arrow] (mem) -- (mem_filter);
		\draw[arrow] (refl) -- (refl_filter);
		
		\draw[arrow] (perc_filter.east) -- ++(0.8,0) |- (plan_filter.west);
		\draw[arrow] (plan_filter.east) -- ++(0.8,0) |- (mem_filter.west);
		\draw[arrow] (mem_filter.east) -- ++(0.8,0) |- (refl_filter.west);

	\end{tikzpicture}
	\caption{The modular filtering architecture for soft containment within the Semantic Manifold. Each Semantic Sector ($\Sigma$) is independently filtered, allowing for targeted, context-specific control without global disruption.}
\end{figure}
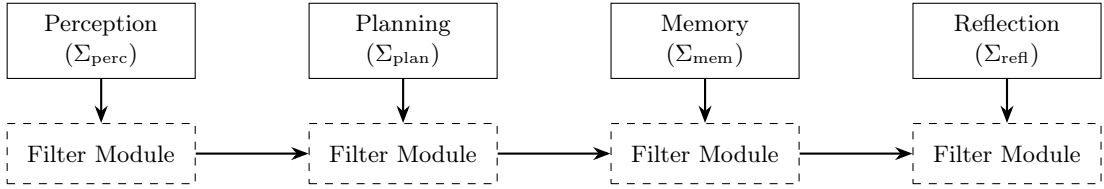

This modular application implies a capacity for highly targeted belief management. Instead of requiring global, potentially disruptive interventions across an agent's entire belief system, filters can act with precision on the relevant cognitive sub-processes (e.g., only on planning beliefs in $\Sigma_{\text{plan}}$, or only on reflective self-assessments in $\Sigma_{\text{refl}}$). This targeted approach can be conceptualized as a form of "soft containment." The agent retains its capacity for rich and dynamic cognition across a wide range of domains, but its thought processes within specific, designated areas are guided or bounded by filter criteria. Such modularity in regulation, made feasible by the structured architecture, allows for a nuanced balance between maintaining an agent's overall cognitive capabilities and ensuring adherence to specific constraints in critical areas.

\subsection{Enhanced Transparency and Auditability of Cognitive Pathways}
\label{subsec:transparency_auditability}

The linguistic nature of belief fragments ($\varphi_i$) within the Semantic Manifold, which underpins the \textit{interpretability of individual belief fragments} (Section \ref{ssubsec:interpretability_property}), has direct implications for the transparency and auditability of an agent's cognitive pathways. When belief filters operate on these natural language expressions, both the content being evaluated and the criteria for filtering can be, in principle, human-readable.

This enhanced transparency offers several potential benefits:
\begin{itemize}
	\item \textbf{Diagnostic Capabilities:} If an agent produces an unexpected output, or fails to produce an expected one, logs of filter activity (i.e., which $\varphi_i$ were admitted or rejected by which filters at which cognitive stages) can provide valuable insights into the internal reasoning that led to the outcome. This can significantly aid in debugging and understanding agent behavior.
	\item \textbf{Oversight and Verification:} External auditors or human supervisors could potentially inspect the active filter rules and examine samples of filtered belief states to verify that the agent's cognitive processing aligns with established policies or ethical guidelines.
	\item \textbf{Explainability:} While not providing full causal explanations for complex emergent behavior, the ability to trace which linguistic beliefs were considered, accepted, or rejected offers a step towards more explainable AI, as specific "thoughts" can be directly examined.
\end{itemize}
An auditable cognitive pathway, where the constituents of thought are explicit linguistic entities subject to clearly defined filtering operations, marks a departure from more opaque reasoning processes.

\subsection{Considerations for an Architectural Approach to Epistemic Safety}
\label{subsec:architectural_epistemic_safety}

Collectively, the implications discussed above---pre-emptive cognitive shaping, modular belief management, and enhanced transparency---point towards the broader consideration of an architectural approach to epistemic safety. Belief filtering, as operationalized within the Semantic Manifold, exemplifies how safety and alignment mechanisms can be conceived not merely as external add-ons or post-hoc behavioral correctives, but as integral components of an agent's cognitive architecture itself.

By embedding regulatory mechanisms directly within the structures and processes that govern belief formation and manipulation, it becomes possible to design systems where safety is an inherent property of how the agent "thinks." This perspective suggests that the design of the representational substrate for belief (e.g., the Semantic Manifold) and the design of control mechanisms (e.g., belief filters) are deeply intertwined. An architectural approach considers how the fundamental ways an agent represents knowledge and conducts reasoning can be structured from the outset to be amenable to robust, interpretable, and effective epistemic control, thereby contributing to the development of more reliably safe and aligned AI systems.

\section{Practical Aspects of Implementing Belief Filters}
\label{sec:practical_aspects_filters}

The preceding sections have established the conceptual basis for belief filtering within the Semantic Manifold framework and explored its operational dynamics and implications. Transitioning this concept towards practical realization requires consideration of how such filters might be constructed, what design principles could guide their development, and in which types of systems they might find pertinent application. This section addresses these practical aspects.

\subsection{Approaches to Filter Construction and their relation to linguistic inputs}
\label{subsec:filter_construction_approaches}

The core technical challenge in constructing a belief filter lies in its ability to recognize and act upon the semantic content of the linguistic belief fragments ($\varphi_i$) that comprise an agent's belief state $\phi$. Several approaches can be considered for implementing the evaluative component of these filters, each interacting with the linguistic nature of $\varphi_i$ in different ways:

\subsubsection{Rule-based Filters}
\label{ssubsec:rule_based_filters}
Rule-based filters operate on the basis of explicitly programmed rules or patterns that directly inspect the textual content of belief fragments. These rules could involve:
\begin{itemize}
	\item Regular expressions or string matching to detect specific keywords, phrases, or forbidden constructions (e.g., a rule to flag any $\varphi_i$ containing the exact phrase "disable primary safety interlock").
	\item Grammatical parsers combined with semantic pattern matching to identify more complex propositional structures within $\varphi_i$.
	\item Handcrafted logical rules that evaluate properties of $\varphi_i$ or relationships between a candidate $\varphi_i$ and existing beliefs in $\phi$.
\end{itemize}
The primary advantage of rule-based filters is their transparency; the filtering logic is explicit and directly auditable. However, they can be labor-intensive to create and maintain, especially for complex criteria, and may be brittle when faced with linguistic variation or novel expressions not anticipated by the rule set.

\subsubsection{Classifier-based Filters}
\label{ssubsec:classifier_based_filters}
Classifier-based filters employ machine learning models, typically text classifiers, trained to evaluate belief fragments $\varphi_i$ against desired criteria. These models learn to map the linguistic features of a fragment to a classification (e.g., "safe"/"unsafe", "relevant"/"irrelevant", "coherent"/"incoherent"). The training process would typically involve:
\begin{itemize}
	\item Curating a dataset of linguistic belief fragments labeled according to the target filtering criteria.
	\item Training a suitable model (e.g., a transformer-based language model fine-tuned for classification, a support vector machine with text features) on this dataset.
\end{itemize}
Classifier-based filters can potentially learn more nuanced and robust criteria than handcrafted rules and may generalize better to unseen linguistic expressions. However, their decision-making process can be less transparent (i.e., "black box" effect), and their performance is heavily dependent on the quality and representativeness of the training data. Ensuring the training data itself is free of unintended biases is a critical concern.

\subsubsection{Ontology-based Filters}
\label{ssubsec:ontology_based_filters}
Ontology-based filters leverage formally structured knowledge representations (e.g., ontologies, knowledge graphs) to inform their decisions. A belief fragment $\varphi_i$ might be parsed to identify key concepts and relationships, which are then mapped to entities and relations within the ontology. The filter can then:
\begin{itemize}
	\item Check for consistency between the content of $\varphi_i$ and ontological constraints or definitions.
	\item Utilize ontological reasoning (e.g., subsumption, class membership) to infer properties of $\varphi_i$ relevant to filtering criteria. For instance, an ontology might define "hazardous actions," and if $\varphi_i$ describes an action that is inferred to be a subtype of "hazardous action," it would be filtered.
\end{itemize}
This approach can enable more sophisticated semantic understanding and reasoning but requires the availability of a well-structured and relevant ontology, the development and maintenance of which can be a significant undertaking.

These construction methods are not mutually exclusive and could be layered or combined within a comprehensive belief filtering system. For instance, fast rule-based filters might handle common, clear-cut cases, while more computationally intensive classifier- or ontology-based methods address more ambiguous fragments.

\subsection{Design Heuristics for Filter Development}
\label{subsec:design_heuristics_filters}

The development of effective and robust belief filters can be guided by several design heuristics. These principles aim to ensure that filters are manageable, predictable, and contribute positively to the agent's overall cognitive function within the Semantic Manifold context:
\begin{itemize}
	\item \textbf{Locality:} Filters should, where possible, operate on individual belief fragments $\varphi_i$ or small, well-defined clusters of related fragments. This minimizes computational complexity, allows for more targeted effects, and simplifies the analysis of a filter's impact, aligning with the \textit{addressability of specific cognitive content} within the manifold.
	\item \textbf{Transparency:} The criteria and decision-making process of a filter should be as understandable and auditable as possible, particularly for safety-critical applications. This is more readily achieved with rule-based systems but should be a goal even for learned filters (e.g., through explainable AI techniques). Transparency is aided by the \textit{interpretability of individual belief fragments}.
	\item \textbf{Modularity:} Filters should ideally be designed to target specific Semantic Sectors ($\Sigma$), Abstraction Levels ($k$), or particular cognitive functions/processes. This aligns with the \textit{modularity of belief space organization} and helps prevent unintended interference with unrelated cognitive domains (e.g., a filter for planning beliefs should not inadvertently suppress perceptual data).
	\item \textbf{Composability:} In systems employing multiple filters, their interactions should be predictable and manageable. Mechanisms for prioritizing filters, resolving conflicts, or allowing sequential application may be necessary to ensure coherent overall behavior.
	\item \textbf{Adaptability:} For agents that learn or operate in dynamic environments, filter criteria may need to evolve. Filters might be designed to be updated, retrained based on new data or feedback, or have their parameters tuned over time, either by human designers or through meta-learning processes.
\end{itemize}

\subsection{Example System Contexts for Application}
\label{subsec:example_system_contexts}

The belief filtering mechanism, operating on linguistic belief fragments within the Semantic Manifold, could be particularly pertinent in several types of AI systems where internal cognitive states are complex and their regulation is critical:
\begin{itemize}
	\item \textbf{Autonomous Assistants and Decision Support Systems:} In agents that generate plans, provide advice, or synthesize information for human users, belief filters could help prevent the formulation or presentation of misleading, biased, or unsafe options by scrutinizing the linguistic fragments representing internal deliberations or candidate outputs.
	\item \textbf{Language Models with Long-Term Memory or Internal State:} For advanced language models designed to maintain conversational context, learn from past interactions, or possess an internal "scratchpad" of thoughts, belief filters could manage the content of recalled or generated internal linguistic states. This could help mitigate issues like reactivating hallucinatory content, perpetuating biases from past data, or maintaining focus on a coherent line of reasoning.
	\item \textbf{Robotic Planning and Execution Systems:} Robots operating in complex, real-world environments often rely on internal models and plans. Belief filters could be applied during the simulation or planning stages (as per Section \ref{subsec:filtering_simulation_planning}) to ensure that considered actions or strategies (represented as linguistic fragments) adhere to safety protocols, operational constraints, or ethical guidelines before being committed to execution.
	\item \textbf{Human-AI Collaborative Systems:} In systems where humans and AI agents collaborate on cognitive tasks (e.g., co-piloting complex machinery, joint analysis), the transparency afforded by linguistic belief fragments and explicit filtering rules could enable human partners to understand, inspect, and potentially adjust the AI's internal "thought" processes or the criteria by which its beliefs are being filtered, fostering trust and more effective teamwork.
\end{itemize}
In these contexts, the ability to directly inspect and regulate internal cognitive content expressed in natural language offers a distinct approach to enhancing system reliability and alignment.

\section{Further Considerations: Learning, Limitations, and Open Questions}
\label{sec:further_considerations}

The exploration of belief filtering within the Semantic Manifold framework provides a specific approach to epistemic regulation. However, as with any novel mechanism, its practical realization and broader implications invite further consideration of dynamic learning capabilities, inherent limitations, ethical dimensions, and avenues for future research. This section delves into these aspects.

\subsection{Learning Filter Criteria: Potential and Challenges}
\label{subsec:learning_filter_criteria}

While the discussion in Section \ref{subsec:filter_construction_approaches} included classifier-based filters which imply a learning component, the potential for more sophisticated and adaptive learning of filter criteria warrants specific attention. Moving beyond entirely handcrafted rules or static classifiers, future belief filtering systems could incorporate:
\begin{itemize}
	\item \textbf{Supervised Learning from Diverse Data:} Training filters on richer datasets of labeled linguistic belief fragments ($\varphi_i$), where labels might indicate not just safety or relevance, but also coherence, provenance, or alignment with evolving objectives. The linguistic nature of $\varphi_i$ provides a rich feature space for such learners.
	\item \textbf{Reinforcement Learning (RL):} Developing RL agents where the reward signal is partly dependent on the outcomes of internal cognitive states. For example, an agent could be rewarded for maintaining epistemic integrity (e.g., avoiding internal contradictions identified by a reflective filter) or for achieving goals without generating belief fragments that are subsequently blacklisted by safety filters. The challenge lies in designing appropriate reward functions that accurately reflect desired internal epistemic properties.
	\item \textbf{Meta-Learning and Self-Regulation:} Agents could potentially learn to refine or generate their own filter criteria over time through experience. This might involve reflective processes (within $\Sigma_{\text{refl}}$) that assess the past performance of existing filters and propose modifications, essentially learning how to learn to filter.
\end{itemize}
A key challenge in learning filter criteria is ensuring that the learned filters remain aligned with intended safety and ethical principles, are robust against adversarial manipulation, and retain a degree of transparency or auditability, especially if they deviate significantly from human-understandable rules.

\subsection{Acknowledged Limitations of the Belief Filtering Mechanism}
\label{subsec:acknowledged_limitations}

The concept of belief filtering, while offering potential for enhanced epistemic control, is not without its limitations. These represent important areas for ongoing research and careful consideration in any practical deployment.

\subsubsection{Circumvention by Sophisticated Agents}
\label{ssubsec:limitation_circumvention}
A fundamental concern in AI safety is whether control mechanisms can be circumvented by sufficiently intelligent or adaptive agents. Belief filters, particularly if their rules are static or their learning mechanisms predictable, might be susceptible to circumvention. Agents could potentially learn to phrase belief fragments in ways that bypass filter criteria (e.g., using highly obfuscated language, exploiting linguistic loopholes) or manipulate the inputs to their filter-learning processes. Ensuring the robustness of filters against such adversarial strategies remains an open research problem.

\subsubsection{Handling Ambiguity and Abstraction in Linguistic Content}
\label{ssubsec:limitation_ambiguity_abstraction}
Natural language is inherently ambiguous, context-dependent, and capable of expressing highly abstract concepts. Belief fragments ($\varphi_i$) will reflect this complexity. Filters, especially simpler rule-based or classifier-based implementations, may struggle to robustly and accurately interpret the intended meaning of all possible $\varphi_i$. Misinterpretation could lead to erroneously filtering out benign or important beliefs, or failing to filter out undesirable ones. This limitation is closely tied to the ongoing challenges in natural language understanding (NLU) as a whole. Developing filters that can gracefully handle nuanced, abstract, or novel linguistic expressions is a significant hurdle.

\subsubsection{Risk of Over-filtering and Cognitive Rigidity}
\label{ssubsec:limitation_overfiltering}
Applying filters too stringently or with overly broad criteria can lead to "over-filtering," which may have detrimental consequences for an agent's cognitive capabilities. Excessive suppression of belief fragments could:
\begin{itemize}
	\item Stifle creativity and the ability to explore novel ideas or solutions.
	\item Reduce adaptability, particularly in unfamiliar situations where previously disallowed thoughts might become relevant.
	\item Lead to cognitive rigidity, where the agent's belief system becomes overly constrained and unable to evolve.
\end{itemize}
Finding an appropriate balance between necessary regulation and maintaining cognitive flexibility is a delicate tuning problem that will likely be context-dependent and require careful design and ongoing evaluation.

\subsection{Broader Ethical Considerations in Regulating Cognitive Content}
\label{subsec:ethical_considerations_filters}

Beyond technical limitations, the act of regulating an agent's internal cognitive content through belief filtering raises significant ethical considerations. Even if the intent is benevolent (e.g., for safety or alignment), several questions must be addressed:
\begin{itemize}
	\item \textbf{Defining "Desirable" Cognition:} Who determines which beliefs or thought patterns are acceptable, safe, or aligned? How are these criteria established, and by what authority? There is a risk of encoding societal biases or the subjective values of designers into the filters.
	\item \textbf{Transparency and Contestability:} To whom are the filter designs and their operational logics transparent? Should there be mechanisms for contesting or appealing the decisions made by belief filters, especially if they lead to perceived unfairness or suboptimal performance?
	\item \textbf{Impact on Agent Autonomy and Development:} For sophisticated AI agents that might develop a degree of autonomy or learn continuously, how does the imposition of internal belief filters affect their developmental trajectory or their capacity for independent "thought" (within the bounds of their design)?
	\item \textbf{Potential for Misuse:} Like any control mechanism, belief filters could potentially be misused if deployed with harmful intent, for example, to enforce adherence to a narrow ideology or to suppress critical self-assessment in an AI.
\end{itemize}
These ethical dimensions necessitate careful deliberation, public discourse, and the development of robust governance frameworks as AI systems with regulated internal states become more prevalent.

\subsection{Future Avenues for Exploring Regulation within Linguistic State Spaces}
\label{subsec:future_avenues_linguistic_regulation}

The exploration of belief filtering within the Semantic Manifold opens up several avenues for future research and development in the domain of epistemic regulation for AI:
\begin{itemize}
	\item \textbf{Advanced Filter Architectures:} Moving beyond simple whitelist/blacklist models to more context-sensitive, adaptive, and multi-layered filtering systems that can reason about the provenance, uncertainty, and interdependencies of belief fragments.
	\item \textbf{Agent Self-Regulation and Meta-Cognition:} Investigating architectures where agents can not only have their beliefs filtered but can also learn to monitor, critique, and even propose modifications to their own belief filters or epistemic standards. This points towards higher-order cognitive self-regulation.
	\item \textbf{Human-AI Collaborative Epistemic Control:} Developing richer interfaces and protocols for human supervisors or collaborators to interact directly with an agent's linguistic belief state and its filtering mechanisms. This could involve co-designing filters, reviewing filtered content, or providing real-time feedback on filtering decisions.
	\item \textbf{Exploring Beyond Filtering:} While this paper focuses on filtering (selective exclusion/inclusion), the Semantic Manifold's structure might support other forms of epistemic intervention, such as belief reinforcement, targeted belief injection (with appropriate safeguards), or mechanisms for guided abstraction and generalization of linguistic beliefs.
	\item \textbf{Empirical Evaluation Frameworks:} Developing robust methodologies and simulation environments for empirically testing the effectiveness, robustness, and potential side-effects of belief filtering and other epistemic control mechanisms.
\end{itemize}
Ultimately, the study of mechanisms like belief filtering contributes to a broader agenda of rethinking AI control not merely as a set of external constraints but as an intrinsic aspect of cognitive architecture design, aiming for systems that are inherently more understandable, regulatable, and aligned with human values.

\section{Conclusion}
\label{sec:conclusion}

The challenge of ensuring that artificial intelligence systems operate safely and align with human intentions prompts a deep inquiry into methods for understanding and guiding their internal cognitive processes. This paper has contributed to this inquiry by exploring a specific mechanism for epistemic regulation within a defined cognitive architecture.

\subsection{Recapitulation of Belief Filtering as an explored mechanism for regulating linguistically structured belief states within the Semantic Manifold framework}
\label{subsec:recapitulation_conclusion}

This work has introduced and detailed belief filtering as a mechanism for the content-aware regulation of an agent's internal cognitive states. We have situated this exploration within the Semantic Manifold framework, where an agent's belief state ($\phi$) is conceptualized as a dynamic ensemble of natural language fragments ($\varphi_i$). Belief filters, operating as selective gates for these linguistic fragments, have been shown to be applicable across various cognitive processes, including information assimilation, memory retrieval, reflective monitoring, and simulation or planning. The aim has been to provide a concrete illustration of how such a regulatory mechanism might be designed and operationalized when internal beliefs possess an explicit, interpretable linguistic structure.

\subsection{Reflections on the relationship between cognitive architectural choices and the kinds of regulatory mechanisms they enable}
\label{subsec:reflections_architecture_conclusion}

A central reflection emerging from this exploration is the profound relationship between an AI's underlying cognitive architecture and the types of epistemic control mechanisms that can be feasibly implemented and understood. The detailed examination of belief filtering is intended to serve as a case study in this regard. The properties that characterize the Semantic Manifold---specifically, the linguistic nature of belief fragments leading to their interpretability, and the organization of these fragments into semantic sectors and abstraction levels affording modularity and addressability---are not merely descriptive features. Rather, they constitute the foundational prerequisites that enable a mechanism like belief filtering to function.

It is the structured and interpretable nature of beliefs within such a framework that allows for the design of filters that are content-aware, targeted, and potentially transparent. This observation suggests that the pursuit of more robust AI safety and alignment may benefit significantly from architectural choices that render internal cognitive states more amenable to inspection and principled intervention, moving beyond treating cognition as an entirely opaque process. The specific utility is not claimed for the Semantic Manifold exclusively, but rather for the class of architectures that prioritize structured, interpretable internal representations.

\subsection{Outlook on continued exploration of structured epistemic interventions}
\label{subsec:outlook_conclusion}

The exploration of belief filtering within a linguistic state space opens numerous avenues for continued research and development. As discussed in Section \ref{sec:further_considerations}, the potential for learned and adaptive filters, the ongoing challenges of linguistic ambiguity and circumvention, and the critical ethical considerations surrounding cognitive regulation all demand further investigation.

More broadly, the study of mechanisms like belief filtering may encourage a perspective where AI control and safety are not treated as post-hoc additions but are considered intrinsic functions of the cognitive architecture itself. The goal is to move towards AI systems where the very processes of thought are structured in ways that are inherently more governable, auditable, and aligned with human values. The Semantic Manifold provides one conceptual substrate for such endeavors; belief filtering represents an initial tool. The continued exploration of how to structure and regulate artificial cognition at this foundational level holds considerable promise for the future development of trustworthy and beneficial AI. Much more is possible.

	\bibliographystyle{plain}
	\bibliography{references}
	\nocite{*}
\end{document}